\documentclass{AISB2008}
\usepackage{newfloat}
\DeclareFloatingEnvironment[fileext=frm,name=Listing]{listing}
\usepackage{placeins}
\usepackage{float}
\usepackage{listings}

\usepackage{times}
\usepackage{graphicx}
\usepackage{latexsym}

\newcommand*{\sourceatright}[1]{\unskip\hspace{1em plus 1fill}%
\nolinebreak[3]\hspace*{\fill}\mbox{#1}}

\usepackage{chngcntr}
\counterwithout{figure}{section}
\counterwithout{figure}{subsection}

\usepackage{amsmath}

\makeatletter
\renewcommand{\boxed}[1]{\text{\fboxsep=.2em\fbox{\m@th$\displaystyle#1$}}}
\makeatother

\newcommand{\mystrut}{\vphantom{b\gamma}}

\usepackage{amsopn}
\newcommand{\droparrow}{%
  \mathchoice{\raisebox{-4pt}{$\displaystyle\mapsto$}}
             {\raisebox{-4pt}{$\mapsto$}}
             {\raisebox{-2pt}{$\scriptstyle\mapsto$}}
             {\raisebox{-2pt}{$\scriptscriptstyle\mapsto$}}}

\usepackage{marvosym,hetcasl}

\begin{document}

\title{The Search for Computational Intelligence}

\author{Joseph Corneli
\institute{Department of Computing, Goldsmiths College, University of London\newline
\Email \url{j.corneli@gold.ac.uk}}
\and
Ewen Maclean
\institute{School of Informatics, University of Edinburgh\newline
\Email \url{ewenmaclean@gmail.com}}}

\maketitle
\bibliographystyle{AISB2008}

\begin{abstract}
We define and explore in simulation several rules for the local
evolution of generative rules for 1D and 2D cellular automata.  Our
implementation uses strategies from conceptual blending.  We discuss
potential applications to modelling social dynamics.
\end{abstract}

\section{Introduction}

This paper takes a local approach to studying the evolution of
cellular automata, following on the global approach of ``PICARD''
\cite{pavlic2014self}.

\begin{quote}
\emph{Like a traditional one-dimensional CA, PICARD executions move
  from one iteration to another by some rule. However, whereas
  traditional CA's require the rule to be static and externally
  specified, PICARD infers the iteration rule from the current state
  of the CA itself.}
\sourceatright{\cite[pp. 1--2]{pavlic2014self}}
\end{quote}

PICARD's inferred rule is derived from the current state of the CA by
a global characteristics, such as the number of 1's in the CA's
current state (modulo 256), or the density $\rho$ of 1's (normalised
as $\rho/256$).  These global criteria are similar to Van Valen's
theory of resource density as an ``incompressible gel'' \cite{van1973new}.

\begin{figure}
\includegraphics[width=\columnwidth]{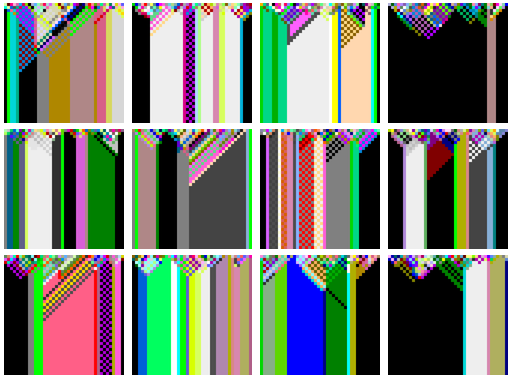}
\caption{An illustration of MetaCA evolution\label{metaca-taster}}
\end{figure}

In the current paper we introduce the notion of a MetaCA, in which CA
rules are derived locally at each cell within the CA as it runs.
Examples appear in Figure \ref{metaca-taster}.  Here, each colour
represents one of the 256 standard one-dimensional CA rules.  States
evolve locally, according to globally-defined dynamics.

\newpage

\section{Background}

\subsection{Cellular Automata}

A crucial development in the history of CA research was the proof
\cite{cook2004universality} that certain CA rules are Turing complete
(in particular, Rule 110 in Wolfram's numbering system
\cite{wolfram1994cellular} enjoys this property).

Earlier classic works
\cite{langton1990computation,mitchell1993revisiting,packard1988adaptation}
exploring related ``edge of chaos'' effects.  In
\cite{packard1988adaptation,mitchell1993revisiting,mitchell1994evolving},
genetic algorithms are used to search the space of CA rules via
crossover and mutation.  This sort of evolution is global and is
connected with the CA rule by a derived parameter, ``Langton's
$\lambda$'' (cf. \cite{langton1990computation}).  An overview of the
``EvCA'' programme is presented in \cite{hordijk2013evca}.

Closest to the work presented here is \cite{sipper1997evolution},
which introduces the paradigm of \emph{cellular programming}.  As the
name indicates, this approach is a fusion of ideas from cellular
automata and genetic programming.
\begin{quote}
\emph{As opposed to the standard genetic algorithm, where a population
  of independent problem solutions globally evolves, our approach
  involves a grid of rules that coevolves locally.}
\sourceatright{\cite[p. 74]{sipper1997evolution}}
\end{quote}
In cellular programming, local evolution of the CA rule makes use of a
local ``fitness'' (\cite[pp. 79--81]{sipper1997evolution}), as the
systems are evolved to perform certain global computational tasks.

In the current effort, although we are interested in behaviour that
tends towards edge-of-chaos effects, system evolution is not directly
guided by a specific fitness criterion, but only by variations on the
``crossover'' mechanism.

One early application of cellular programming was to evolutionary game
theory, a field with natural parallels
(cf.~\cite{nowak1992evolutionary}).  We will not consider game
theoretic approaches in this work, despite being inspired by the
social metaphors that are involved (e.g.~\cite{nowak2006five}).  In
our thinking we often switch between conceptual/symbolic,
social/ethical, biological/genetic, and physical/geometric metaphors.

In this connection it is worth mentioning some recent work
\cite{goerg2012licors,goerg2012mixed} that continues in the earlier
tradition of the EvCA project (cf. \cite{hordijk2001upper}), making
use of a relativistic ``light cone'' analysis to identify structure in
CAs.  The current paper does not pursue any detailed \emph{post hoc}
analysis of CA behaviour, although we plan to explore this further in
subsequent work.  Finally, although not focusing on CAs per se,
\cite{hofstadter1995prolegomena} outlines a set of criteria for the
design of systems that exhibit ``emergent'' intelligence which helped
to motivate the present effort.

\subsection{Conceptual Blending} \label{sec:blend}

One of our inspirations for working with cellular automata is that we
are involved with a research project that studies computational
blending \cite{schorlemmer2014coinvent}, and cellular automata seem to
offer a very simple example of blending behaviour.  That is, they
consider the value of neighbouring cells, and produce a result that
``combines'' these results (in some suitably abstract sense) in order
to produce the next generation.  We were also inspired by the idea of
``blending'' ordered and chaotic behaviour to produce edge-of-chaos
effects.

We propose to exploit existing formalisms of blending (in the style of
Goguen \cite{gog05}) in the context of cellular automata to
investigate emergent and novel behaviours.  The fundamental building
blocks used in calculating concept or theory blends are:
\begin{description}
\item[Input Concepts] are the concepts or theories which are understood have some degree of commonality (syntactic or semantic). 
\item[Signature Morphism] is a definition of how symbols are mapped between theories or concepts. 
\item[Generic Space] is the space which contains a theory which is common to both input theories.
\item[Blend] is the space computed by combining both theories. The computation is computed using a ``pushout'' from the underlying categorical semantics \cite{MossakowskiEA06}. 
\end{description}

Once a blend has been computed, it may represent a concept which is in
some way inconsistent. Equally it may represent a concept which is in
some way incomplete. We can then either weaken an input theory, or
refine the blend:
\begin{description}
\item[Weakening] Given an inconsistent blend it is possible to weaken the input concept in order to produce a consistent blend. Weakening means removing symbols or axioms from the input concept.
\item[Refinement] Given a blend which represents a concept which is in some way incomplete, it is possible to refine the concept by adding symbols or axioms.
\end{description}

This paper presents several examples of simple concepts to which the
blending process applies. In general the notion of a signature
morphism allows input concepts expressed in different languages to be
blended. In this paper the examples shown have input concepts
expressed in the same language, and indeed have the same
specification. This means that the morphisms are not interesting and
the calculated pushout could be computed without utilising the full
machinery of category theory.  Planned extensions will explore the
idea of combining rules for cellular automata which may have entirely
different techniques for expressing propagation.  For this reason, we
target the Heterogeneous Tool Set (HETS) system
\cite{mossakowski2007heterogeneous} as an infrastructure for computing
blends.  We describe our current approach to blending in the context
of cellular automata in Sections \ref{introducing-blending} and
\ref{2d-experiments-design}.


\section{Implementation}

\subsection{Generating Genotypes} \label{sec:geno}

Each elementary CA rule defines a mapping from all eight strings of
0's and 1's to the set \{0,1\}.  Thus, for example the rule \textbf{01010100}
is defined as the following operation:
\begin{lstlisting}[mathescape]
0 0 0 $\mapsto$ $\mathbf{0}$
0 0 1 $\mapsto$ $\mathbf{1}$
0 1 0 $\mapsto$ $\mathbf{0}$
0 1 1 $\mapsto$ $\mathbf{1}$
1 0 0 $\mapsto$ $\mathbf{0}$
1 0 1 $\mapsto$ $\mathbf{1}$
1 1 0 $\mapsto$ $\mathbf{0}$
1 1 1 $\mapsto$ $\mathbf{0}$
\end{lstlisting}

There are 256 of these rules; the example above is Rule 84 in
Wolfram's standard enumeration of 1D CAs \cite{wolfram1994cellular}.
The basic concept of the MetaCA is to evolve a CA with 256 possible
states -- rather than the traditional two -- where each state now
corresponds to a ``CA rule''.  Then we can then apply this rule to
decide the output for the next cell, depending also on the state of
the neighbouring cells.  By positioning three CA rules next to each
other, we define a multiplication by applying the central rule bitwise
across the alleles.
For example, here is the result of ``multiplying'' $01101110\times
01010100\times 01010101$.  In the context of such an operation, we
refer to the central term as the ``local rule,'' and we highlight it
in bold below.

\lstset{
  xleftmargin=.1\columnwidth, xrightmargin=.01\columnwidth
}

\begin{lstlisting}[mathescape]
0 $\mathbf{0}$ 0     $0$     $\text{\emph{Apply local rule to ``000''}}$
1 $\mathbf{1}$ 1     $0$     $\text{\emph{Apply local rule to ``111}}$
1 $\mathbf{0}$ 0     $0$     $\text{\emph{Apply local rule to ``100''}}$
0 $\mathbf{1}$ 1  $\droparrow$   $1$     $\text{\emph{Apply local rule to ``011''}}$
1 $\mathbf{0}$ 0     $0$     $\text{\emph{Apply local rule to ``100''}}$
1 $\mathbf{1}$ 1     $0$     $\text{\emph{Apply local rule to ``111''}}$
1 $\mathbf{0}$ 0     $0$     $\text{\emph{Apply local rule to ``100''}}$
0 $\mathbf{0}$ 1     $1$     $\text{\emph{Apply local rule to ``001''}}$
\end{lstlisting}

\begin{figure}
\includegraphics[width=\columnwidth,trim = 135mm 177mm 0mm 0mm,clip=true]{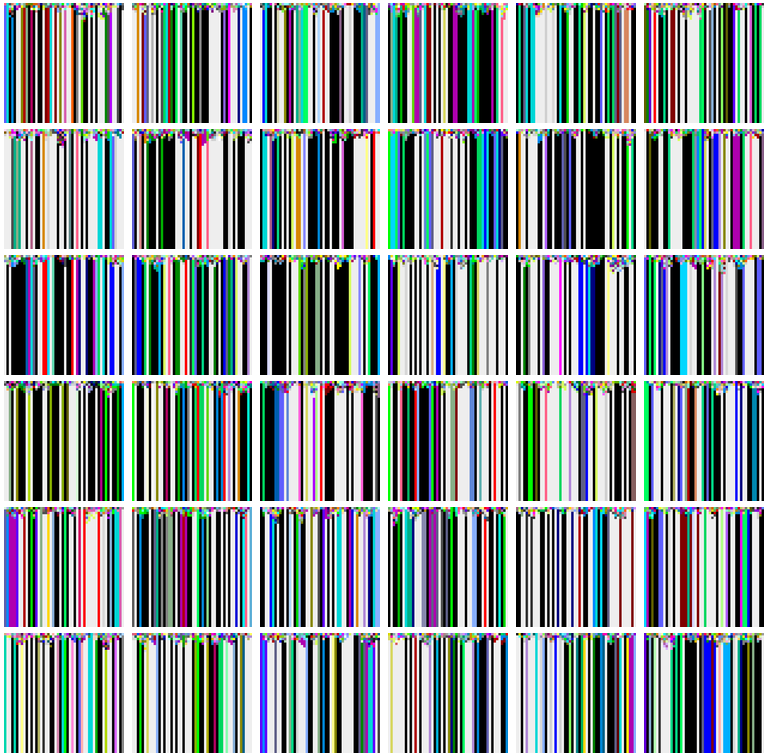}
\caption{Under evolution according to the local rule without blending
  dynamics, a barcode-like stable pattern forms
  quickly\label{barcode}}
\end{figure}

Realised in a simulation with random starting conditions, the results
of this operation are not particularly impressive: they stabilise
early and do not produce any interesting patterns (Figure
\ref{barcode}).

\subsection{Introducing Blending} \label{introducing-blending}


The blending variant says to first compute the ``generic space'' by
noting the alleles where the two adjacent neighbours are the same, and
where they differ.  Only when the generic space retains some ambiguity
(indicated by $\{0,1\}$) do we apply the local rule (again recorded
on the centre cell at left and highlighted in bold) in a bitwise
manner across each allele, to arrive at the final result.

\lstset{
  xleftmargin=.05\columnwidth, xrightmargin=.01\columnwidth
}

\begin{lstlisting}[mathescape]
0 $\mathbf{0}$ 0       0       $\:0$    $\text{\emph{Neighbours are both 0}}$
1 $\mathbf{1}$ 1       1       $\:1$    $\text{\emph{Neighbours are both 1}}$
1 $\mathbf{0}$ 0     {0,1}     $\boxed{0}$    $\text{\emph{Apply local rule to ``100''}}$
0 $\mathbf{1}$ 1  $\droparrow$   {0,1}  $\droparrow$   $\boxed{1}$    $\text{\emph{Apply local rule to ``011''}}$
1 $\mathbf{0}$ 0     {0,1}     $\boxed{0}$    $\text{\emph{Apply local rule to ``100''}}$
1 $\mathbf{1}$ 1       1       $\:1$    $\text{\emph{Neighbours are both 1}}$
1 $\mathbf{0}$ 0     {0,1}     $\boxed{0}$    $\text{\emph{Apply local rule to ``100''}}$
0 $\mathbf{0}$ 1     {0,1}     $\boxed{1}$    $\text{\emph{Apply local rule to ``001''}}$
\end{lstlisting}

\begin{figure}[!ht]
\includegraphics[width=0.45\textwidth]{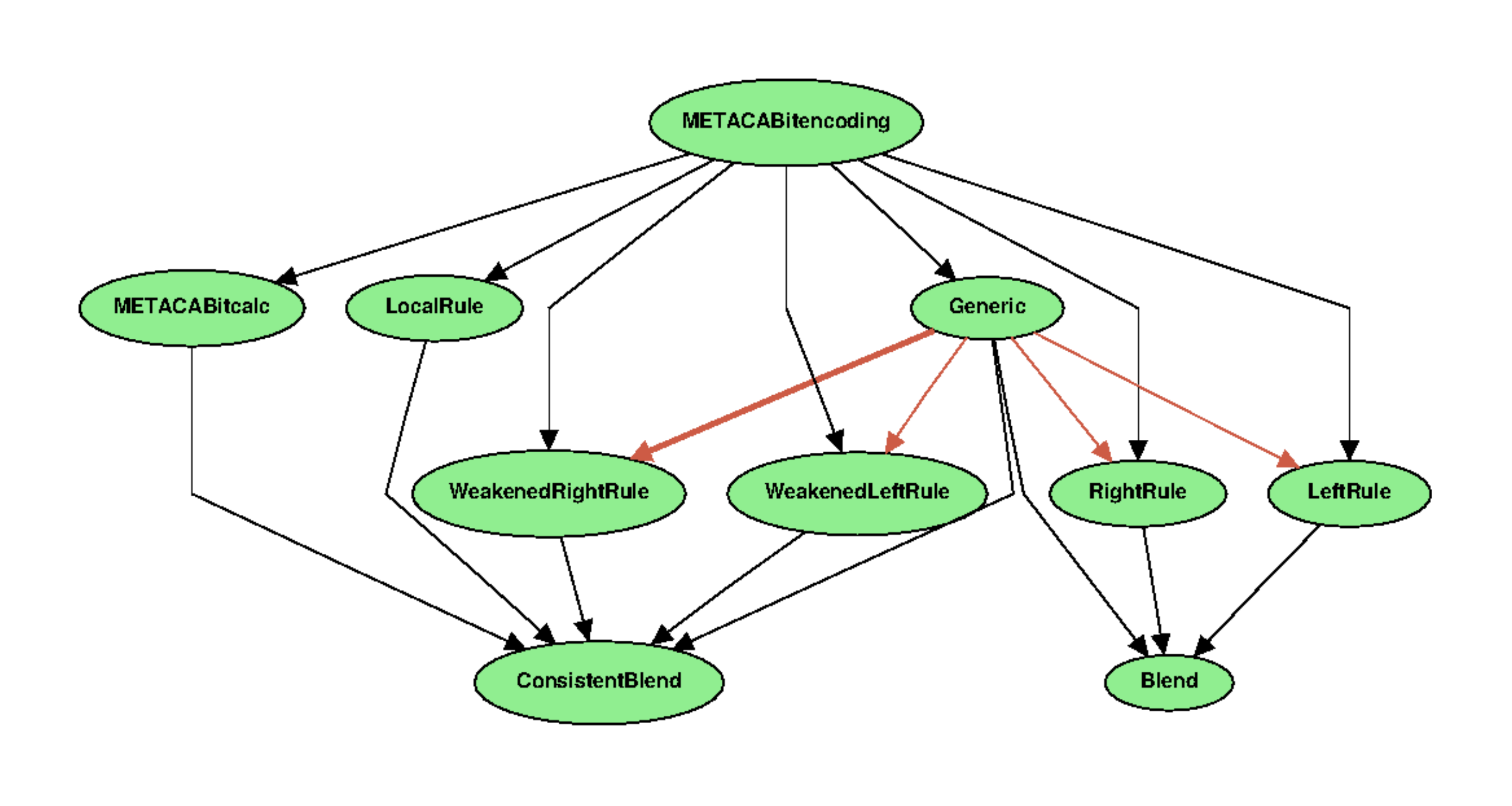}
\caption{The development graph for calculating a blend of 8 bit encodings\label{fig:hetsblend}}
\end{figure}

For illustrative purposes, this blend has been formalised in the HETS
system by introducing CASL files to represent the 8 bit encodings
(Listing \ref{CASL-listing}, and corresponding development graph shown
in Figure \ref{fig:hetsblend}).

The computed blend is inconsistent as there is not a unique value
representing the output value of each function.  In order to resolve
this we weaken the input rules in CASL by removing the function values
which cause conflict. 
Note that purposes of efficiency, we have implemented our 1D
experiments in LISP rather than in \mbox{HETS}/\mbox{CASL}.  We've put
the working code on
Github\footnote{\url{https://github.com/holtzermann17/metaca}}.

\subsection{2D Experiments} \label{2d-experiments-design}



In order to extend the ideas presented so far in the 1D case, let us
consider a variant of Conway's Game of Life \cite{conway}, in which a
global rule exists defining whether a square is alive or dead.  We
extend this by introducing the notion of a local rule at each square
-- a genotype, which governs the propagation of the phenotype.

In Conway's Game of life, one can view the rules for propagation as
partitions on a finite interval $[0,8]$.
\begin{center}
\includegraphics[width=0.45\textwidth]{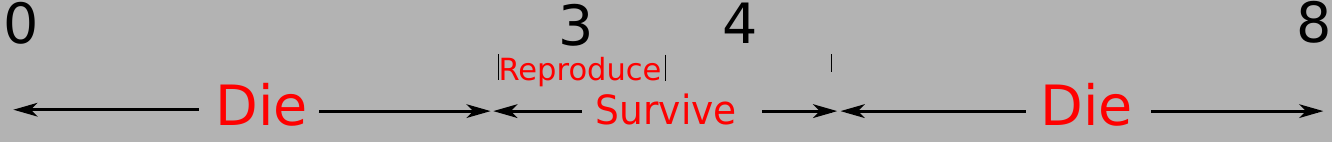}
\end{center}
The number on the line corresponds to the number of alive neighbours
adjacent, in cardinal and inter-cardinal directions, to a given
square. If the square is dead then it becomes alive (labelled
reproduce) if the number of alive neighbours is exactly three. If
there are five or more alive neighbours the square dies from
overcrowding. If there are fewer than three alive neighbours the
square dies from underpopulation. In all other cases the square
maintains its status.


This can be generalised to partitions within a more finely grained
line, for example from 0 to 1000, one creates a genotype $(x,y,z)$:
\begin{center}
\includegraphics[width=0.45\textwidth]{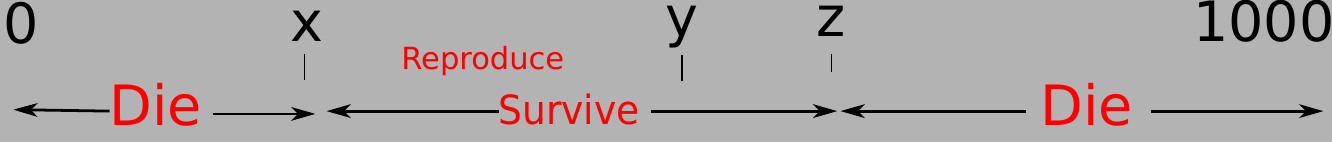}
\end{center}
We introduce the corresponding notion of a {\em weight} for each cell.
The {\em phenotype} of the cell is then a pair $(\text{\emph{alive}},
\text{\emph{weight}})$ which denotes whether the cell is alive, and
what weight is has.  In this paper we always calculate a newly
propagated weight as the average of the neighbours' weights.

The notion of local propagation is introduced by allowing the
genotypes to be blended at each point where a cell remains or becomes
alive.  As we have represented the genotype as a partitioned line, we
can, for example perform a blend where the partition is blended in
such a way as the minimise the lowest bound and maximise the highest
bound, and maximise the interval for reproduction.  Given two
genotypes $(x_1, y_1, z_1)$ and $(x_2, y_2, z_2)$, the blend is
$(\mathrm{min} \{x_1,x_2\}, \mathrm{max} \{y_1,y_2\}, \mathrm{max}
\{z_1,z_2\})$
\begin{center}
\includegraphics[width=0.45\textwidth]{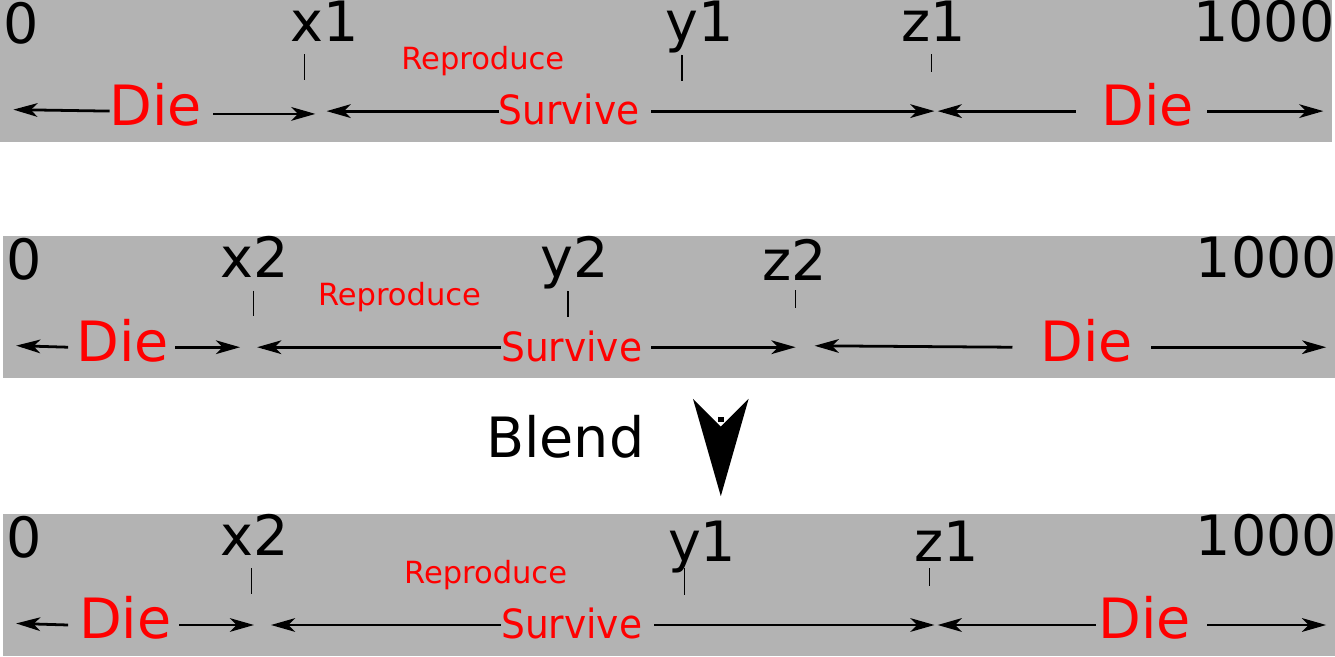}
\end{center}

Note that this is just one of several possible blending strategies,
which we refer to as a {\em union} blend, since it maximises the
partitions which pertain to survival.  We consider alternative blends
in our experiments.



\section{Results}
\subsection{1D CAs} \label{1d-results}

\begin{figure}
\includegraphics[width=\columnwidth]{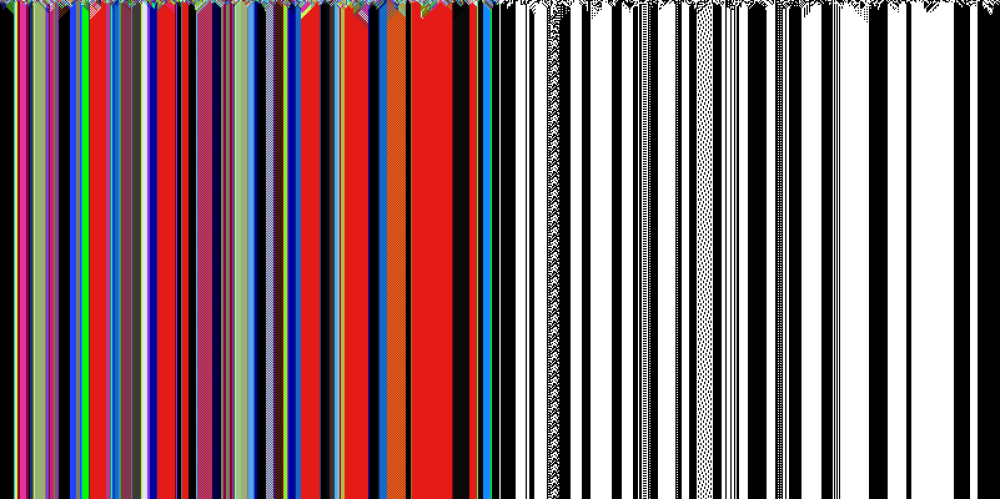}
\caption{Phenotype with behaviour determined by genotype\label{flag}}
\end{figure}

One of the first things we noticed was that even though the blending
dynamic creates more interesting ``CA-like'' patterns than simple
evolution according to the local rule (as illustrated in Figure
\ref{metaca-taster}), it also forms stable bands after this
interesting initial period.  In Figure \ref{flag}, this is illustrated
in a CA running with 500 cells over 500 generations.  Figure
\ref{flag} also includes a phenotype (in black and white) which is
driven entirely by the genotype: that is, if the local genotype is
\boxed{\alpha\mystrut}\boxed{\beta\mystrut}\boxed{\gamma\mystrut} 
where $\alpha, \beta, \gamma \in \{0,1\}^8$
and the local phenotype is
\boxed{a\mystrut}\boxed{b\mystrut}\boxed{c\mystrut}
where $a, b, c \in \{0,1\}$,
then the genotype evolves locally according to the meta-rule $\alpha
\times \beta \times \gamma$ (in the blending variant) while the
phenotype evolves by applying the local rule $\beta$ to the data
``$abc$.''

In the phenotype layer, we see a few bands with interesting patterns,
where the MetaCA at left has stabilised locally into one of the more
interesting CA rules.  However, the long term evolution is not
particularly interesting: the structure observed in Figure
\ref{metaca-taster} disappears quickly.

\begin{figure}
\includegraphics[width=\columnwidth]{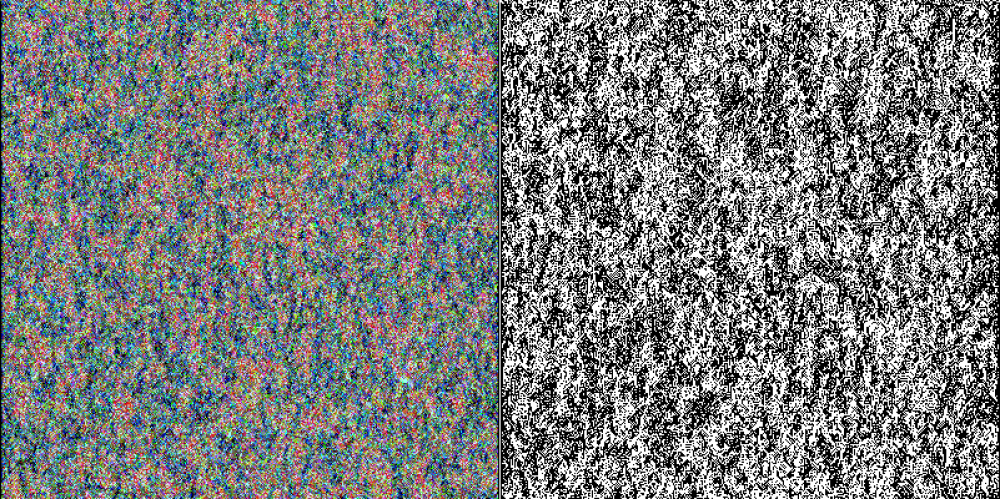}
\caption{A high rate of mutation produces tantalising random structures \label{random-mutation}}
\end{figure}

\begin{figure}
\includegraphics[width=\columnwidth]{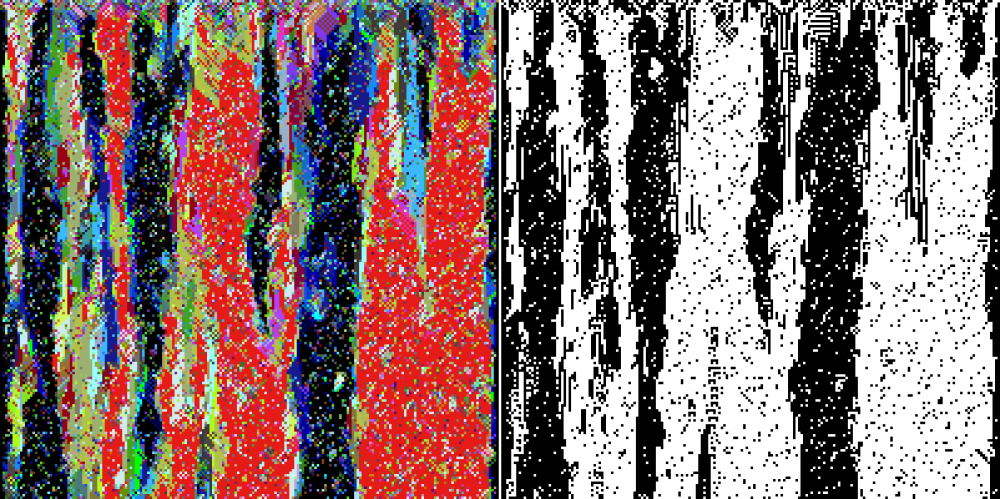}
\caption{Throttling down the mutation rate preserves some of the large-scale stability while making room for variability \label{lower-rate}}
\end{figure}

\begin{figure}
\includegraphics[width=\columnwidth]{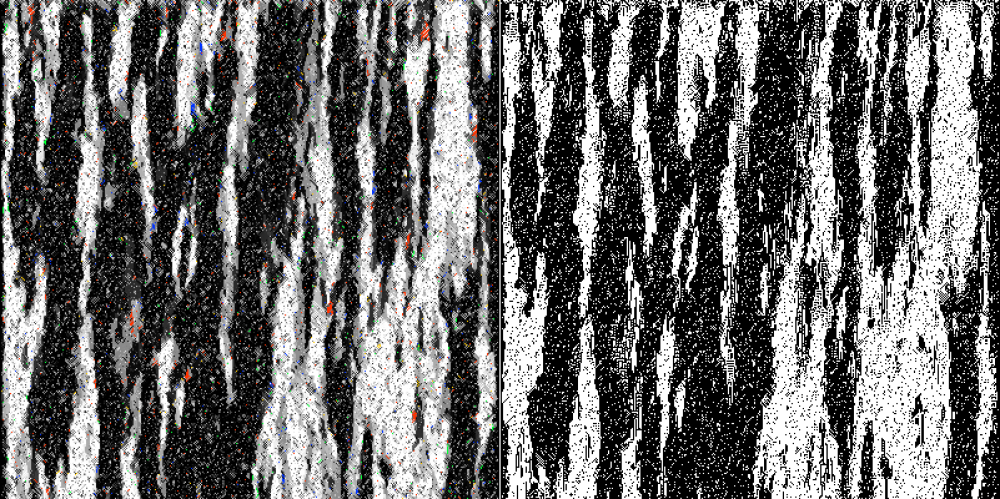}
\caption{The search for intelligent life in the computational universe \label{seti}}
\end{figure}

We therefore decided to introduce random mutations to the genotype,
illustrated in Figures \ref{random-mutation}--\ref{seti}.  With a high
mutation rate, both genotype and phenotype are almost reduced to
confetti.  If we reduce the mutation rate sufficiently, some degree of
stability is preserved, and the vertically striped bands are
transformed into intermingling swaths of colour (Figure
\ref{lower-rate}).  We also see areas with more finely-grained
structure in the phenotype layer.  

In Figure \ref{seti}, the colour-coded genotype layer has been
replaced with a greyscale coding, and we see more clearly how the
phenotype behaviour follows that of the genotype.  That is, genotypes
similar to Rule 0 (00000000) or Rule 256 (11111111) tend to produce 0
or 1, respectively, in the phenotype layer.  Rules that output a blend
of 0's and 1's are mapped to grey shades.  Several interesting rules
(Rule 110, Rule 30, Rule 90, Rule 184, and their reversals, bitwise
inverses, and inverted-reversals) are highlighted in colour.  In
particular, Rule 110 variants are highlighted in red.


We observe that Rule 0 and Rule 256 behaviour tends to predominate.
Grey areas appear to be semi-stable.  Red patches appear and
disappear, as if independent planets evolve intelligent life and are
then extinguished.  With this physics, ``intelligent life'' seems
inevitable, but also inevitably short-lived.  One would have to look
for another overall physics for intelligent behaviour to predominate.

\begin{figure}
\includegraphics[width=\columnwidth]{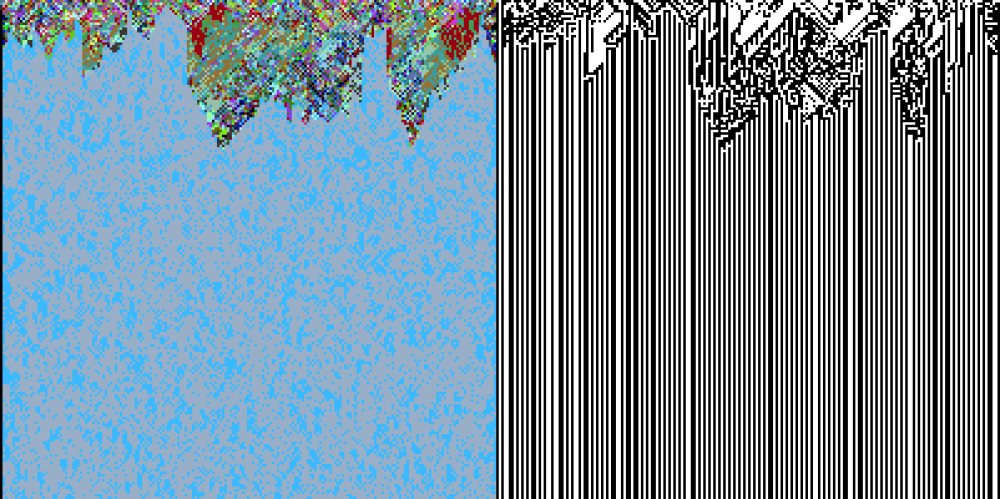} \newline
\includegraphics[width=\columnwidth]{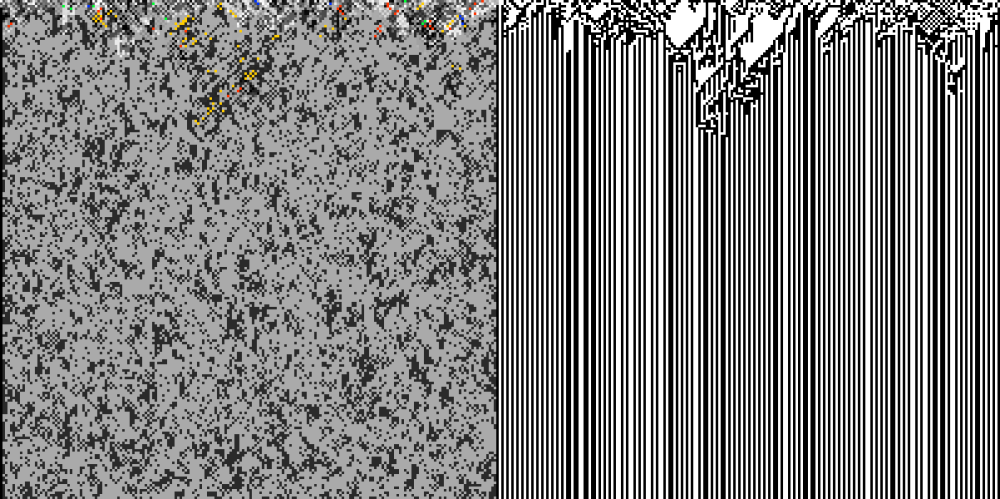} 
\caption{A skewed mutation pattern \label{reef}}
\end{figure}

A potential indication of the direction to look in is presented in
Figure \ref{reef}, which presents CAs generated by adjusting the
typical blending evolution pattern by an (erroneously-programmed)
mutation rule that only flips the first bit.  We see that long-term
behaviour in the genotype flutters randomly between Rule 0 (00000000)
and Rule 128 (10000000).  The short-term behaviour in the phenotype is
nevertheless quite interesting, exhibiting many of the familiar
lifelike edge-of-chaos patterns before ultimately succumbing to a
version of Newton's First Law.

\subsection{2D CAs} \label{2d-results}

To see the behaviour of the union blend in action consider an
initially populated grid, where colours represent the weights of alive
cells:
 \begin{center}
 \includegraphics[width=0.45\textwidth]{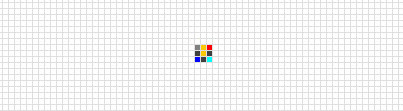}
 \end{center}
For this example, we initially restrict the computation of the blend for a particular cell to take place when the cell is alive in the next iteration. Also we compute the blend of genotype for all neighbours, whether dead or alive.

After 300 iterations the colony has grown a small amount:
 \begin{center}
 \includegraphics[width=0.45\textwidth]{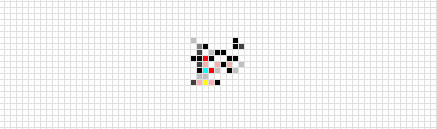}
 \end{center}
Over time, the population continues to grow, with large patches of low-weight (black) cells:
 \begin{center}
 \includegraphics[width=0.45\textwidth]{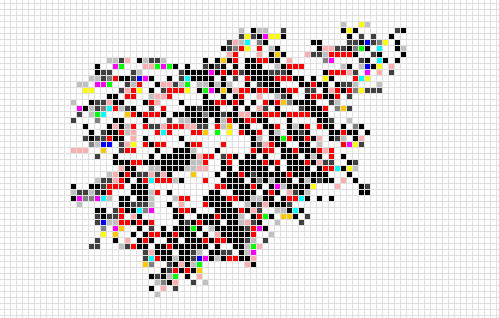}
 \end{center}
Finally some structure starts to appear in the clustering:
 \begin{center}
 \includegraphics[width=0.45\textwidth]{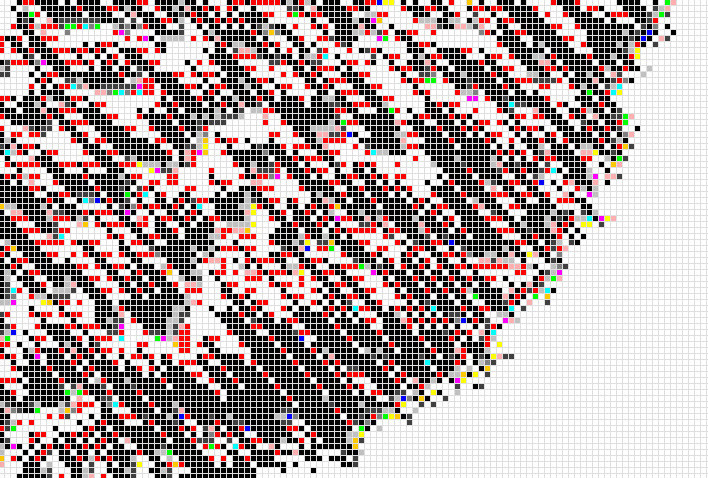}
 \end{center}
The propagation that follows shows a population of cells which grows
slowly overtime. The majority of the members have low weight
(represented by black squares), but interspersed within the population
are chains of squares with high weight (represented by red squares)
adjacent to dead cells (white).

\subsubsection{Modified Blends}

So far we have only showed the union blend working on the genotype. However, it is possible to use different blending techniques:
\begin{itemize}
\item{Consider blending only the genotypes of alive neighbours, or all neighbours;}
\item{Consider only blending genotypes for cells which are alive after propagation;}
\item{Consider an \emph{intersection} blend, where the partition sizes for survival are minimised;}
\item{Consider an \emph{average} blend, where the values of each genotype $(x_i,y_i,z_i)$ are summed and divided by either the number of alive neighbours, or the total number of neighbours.}
\end{itemize}
As an example of different observed emergent behaviour consider a union blend where the blend is only computed from alive neighbours, and as before we compute only for cells which are alive at the next iteration. We start with an initial state:
 \begin{center}
 \includegraphics[width=0.45\textwidth]{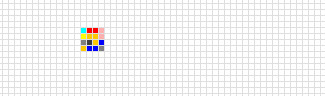}
 \end{center}
and observe a changing, but relatively steady pattern (resembling the motion of a flame) which does not grow in size using the union blend:
 \begin{center}
 \includegraphics[width=0.45\textwidth]{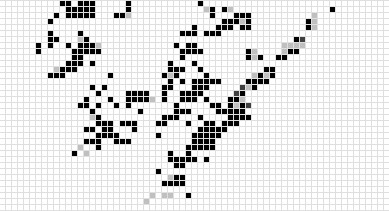}
 \end{center}
\noindent where the weight characteristic of the phenotype of each
cell has fallen to very low.

Finally, consider applying instead an average blend under the same
initial conditions:
 \begin{center}
 \includegraphics[width=0.45\textwidth]{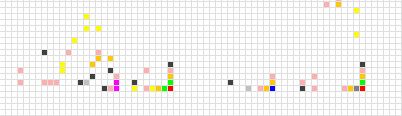}
 \end{center}
Then we see a less steady but more active growth, with populations moving in triangular shapes away from population centres, leaving very small but steady and inactive populations behind:
 \begin{center}
 \includegraphics[width=0.45\textwidth]{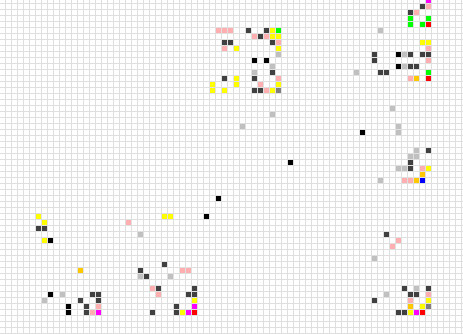}
 \end{center}
\noindent the quickly-moving populations do not have a convergent weight characteristic of the phenotype, as in the case with the union blend for the same initial conditions.


\FloatBarrier

\begin{listing}[H]
{\scriptsize
\begin{hetcasl}
\KW{library} \Id{metaca}\\
\\
\KW{logic} \SId{CASL}\\
\\
\SPEC \=\SIdIndex{METACABitencoding} \Ax{=}\\
\> \KW{free} \KW{type} \=\Id{Bit} \Ax{:}\Ax{:}\=\Ax{=} \Ax{0} \AltBar{} \Ax{1}\\
\> \SORT \Id{Triple}\\
\> \OPS \=\Id{t} \Ax{:} \=\Id{Bit} \Ax{\times} \Id{Bit} \Ax{\times} \Id{Bit} \Ax{\rightarrow} \Id{Triple};\\
\>\> \Id{bitop}\Ax{\_\_} \Ax{:} \=\Id{Triple} \Ax{\rightarrow} \Id{Bit}\\
\KW{end}\\
\\
\%\% How to calculate a blend given three 8-bit genotypes\\
\SPEC \=\SIdIndex{METACABitcalc} \Ax{=}\\
\> \SId{METACABitencoding}\\
\THEN \=\OP \=\Id{blend}\Ax{\_\_}\Ax{\_\_} \Ax{:} \=\Id{Triple} \Ax{\times} \Id{Triple} \Ax{\rightarrow} \Id{Bit}\\
\> \Ax{\forall} \=\Id{t1}, \Id{t2}, \Id{t3} \Ax{:} \Id{Triple} \\
\> \Ax{\bullet} \=\Id{bitop} \Id{t1} \Ax{=} \Id{bitop} \Id{t2} \Ax{\Rightarrow} \=\Id{blend} \Id{t1} \Id{t2} \Ax{=} \Id{bitop} \Id{t1}\\
\> \Ax{\bullet} \=\Ax{\neg} \=\Id{bitop} \Id{t1} \Ax{=} \Id{bitop} \Id{t2} \Ax{\Rightarrow} \=\Id{blend} \Id{t1} \Id{t2} \Ax{=} \Id{bitop} \Id{t3}\\
\KW{end}\\
\\
\SPEC \=\SIdIndex{LeftRule} \Ax{=}\\
\> \SId{METACABitencoding}\\
\THEN \=\Ax{\bullet} \=\Id{bitop} \Id{t}(\=\Ax{0}, \Ax{0}, \Ax{0}) \Ax{=} \Ax{0}\\
\> \Ax{\bullet} \=\Id{bitop} \Id{t}(\=\Ax{0}, \Ax{0}, \Ax{1}) \Ax{=} \Ax{1}\\
\> \Ax{\bullet} \=\Id{bitop} \Id{t}(\=\Ax{0}, \Ax{1}, \Ax{0}) \Ax{=} \Ax{1}\\
\> \Ax{\bullet} \=\Id{bitop} \Id{t}(\=\Ax{0}, \Ax{1}, \Ax{1}) \Ax{=} \Ax{0}\\
\> \Ax{\bullet} \=\Id{bitop} \Id{t}(\=\Ax{1}, \Ax{0}, \Ax{0}) \Ax{=} \Ax{1}\\
\> \Ax{\bullet} \=\Id{bitop} \Id{t}(\=\Ax{1}, \Ax{0}, \Ax{1}) \Ax{=} \Ax{1}\\
\> \Ax{\bullet} \=\Id{bitop} \Id{t}(\=\Ax{1}, \Ax{1}, \Ax{0}) \Ax{=} \Ax{1}\\
\> \Ax{\bullet} \=\Id{bitop} \Id{t}(\=\Ax{1}, \Ax{1}, \Ax{1}) \Ax{=} \Ax{0}\\
\KW{end}\\
\\
\SPEC \=\SIdIndex{RightRule} \Ax{=}\\
\> \SId{METACABitencoding}\\
\THEN \=\Ax{\bullet} \=\Id{bitop} \Id{t}(\=\Ax{0}, \Ax{0}, \Ax{0}) \Ax{=} \Ax{0}\\
\> \Ax{\bullet} \=\Id{bitop} \Id{t}(\=\Ax{0}, \Ax{0}, \Ax{1}) \Ax{=} \Ax{1}\\
\> \Ax{\bullet} \=\Id{bitop} \Id{t}(\=\Ax{0}, \Ax{1}, \Ax{0}) \Ax{=} \Ax{0}\\
\> \Ax{\bullet} \=\Id{bitop} \Id{t}(\=\Ax{0}, \Ax{1}, \Ax{1}) \Ax{=} \Ax{1}\\
\> \Ax{\bullet} \=\Id{bitop} \Id{t}(\=\Ax{1}, \Ax{0}, \Ax{0}) \Ax{=} \Ax{0}\\
\> \Ax{\bullet} \=\Id{bitop} \Id{t}(\=\Ax{1}, \Ax{0}, \Ax{1}) \Ax{=} \Ax{1}\\
\> \Ax{\bullet} \=\Id{bitop} \Id{t}(\=\Ax{1}, \Ax{1}, \Ax{0}) \Ax{=} \Ax{0}\\
\> \Ax{\bullet} \=\Id{bitop} \Id{t}(\=\Ax{1}, \Ax{1}, \Ax{1}) \Ax{=} \Ax{1}\\
\KW{end}\\
\\
\SPEC \=\SIdIndex{LocalRule} \Ax{=}\\
\> \SId{METACABitencoding}\\
\THEN \=\Ax{\bullet} \=\Id{bitop} \Id{t}(\=\Ax{0}, \Ax{0}, \Ax{0}) \Ax{=} \Ax{0}\\
\> \Ax{\bullet} \=\Id{bitop} \Id{t}(\=\Ax{0}, \Ax{0}, \Ax{1}) \Ax{=} \Ax{1}\\
\> \Ax{\bullet} \=\Id{bitop} \Id{t}(\=\Ax{0}, \Ax{1}, \Ax{0}) \Ax{=} \Ax{0}\\
\> \Ax{\bullet} \=\Id{bitop} \Id{t}(\=\Ax{0}, \Ax{1}, \Ax{1}) \Ax{=} \Ax{1}\\
\> \Ax{\bullet} \=\Id{bitop} \Id{t}(\=\Ax{1}, \Ax{0}, \Ax{0}) \Ax{=} \Ax{0}\\
\> \Ax{\bullet} \=\Id{bitop} \Id{t}(\=\Ax{1}, \Ax{0}, \Ax{1}) \Ax{=} \Ax{1}\\
\> \Ax{\bullet} \=\Id{bitop} \Id{t}(\=\Ax{1}, \Ax{1}, \Ax{0}) \Ax{=} \Ax{0}\\
\> \Ax{\bullet} \=\Id{bitop} \Id{t}(\=\Ax{1}, \Ax{1}, \Ax{1}) \Ax{=} \Ax{0}\\
\KW{end}\\
\\
\%\% Generic is common between left and right\\
\SPEC \=\SIdIndex{Generic} \Ax{=}\\
\> \SId{METACABitencoding}\\
\THEN \=\Ax{\bullet} \=\Id{bitop} \Id{t}(\=\Ax{0}, \Ax{0}, \Ax{0}) \Ax{=} \Ax{0}\\
\> \Ax{\bullet} \=\Id{bitop} \Id{t}(\=\Ax{0}, \Ax{0}, \Ax{1}) \Ax{=} \Ax{1}\\
\> \Ax{\bullet} \=\Id{bitop} \Id{t}(\=\Ax{1}, \Ax{0}, \Ax{1}) \Ax{=} \Ax{1}\\
\KW{end}\\
\\
\%\% Morphism from Generic to Left\\
\VIEW \=\SId{Left} \Ax{:} \=\SId{Generic} \KW{to} \SId{LeftRule}\\
\KW{end}\\
\\
\%\% Morphism from Generic to Right\\
\VIEW \=\SId{Right} \Ax{:} \=\SId{Generic} \KW{to} \SId{RightRule}\\
\KW{end}\\
\\
\%\% This will be inconsistent\\
\SPEC \=\SIdIndex{Blend} \Ax{=}\\
\> \KW{combine} \=\Id{Left}, \Id{Right}\\
\KW{end}\\
\\
\SPEC \=\SIdIndex{WeakenedLeftRule} \Ax{=}\\
\> \SId{METACABitencoding}\\
\THEN \=\Ax{\bullet} \=\Id{bitop} \Id{t}(\=\Ax{0}, \Ax{0}, \Ax{0}) \Ax{=} \Ax{0}\\
\> \Ax{\bullet} \=\Id{bitop} \Id{t}(\=\Ax{0}, \Ax{0}, \Ax{1}) \Ax{=} \Ax{1}\\
\> \Ax{\bullet} \=\Id{bitop} \Id{t}(\=\Ax{0}, \Ax{1}, \Ax{0}) \Ax{=} \Ax{1}\\
\> \Ax{\bullet} \=\Id{bitop} \Id{t}(\=\Ax{0}, \Ax{1}, \Ax{1}) \Ax{=} \Ax{0}\\
\> \Ax{\bullet} \=\Id{bitop} \Id{t}(\=\Ax{1}, \Ax{0}, \Ax{0}) \Ax{=} \Ax{1}\\
\> \Ax{\bullet} \=\Id{bitop} \Id{t}(\=\Ax{1}, \Ax{0}, \Ax{1}) \Ax{=} \Ax{1}\\
\> \Ax{\bullet} \=\Id{bitop} \Id{t}(\=\Ax{1}, \Ax{1}, \Ax{0}) \Ax{=} \Ax{1}\\
\KW{end}\\
\\
\SPEC \=\SIdIndex{WeakenedRightRule} \Ax{=}\\
\> \SId{METACABitencoding}\\
\THEN \=\Ax{\bullet} \=\Id{bitop} \Id{t}(\=\Ax{0}, \Ax{0}, \Ax{0}) \Ax{=} \Ax{0}\\
\> \Ax{\bullet} \=\Id{bitop} \Id{t}(\=\Ax{0}, \Ax{0}, \Ax{1}) \Ax{=} \Ax{1}\\
\> \Ax{\bullet} \=\Id{bitop} \Id{t}(\=\Ax{1}, \Ax{0}, \Ax{1}) \Ax{=} \Ax{1}\\
\> \Ax{\bullet} \=\Id{bitop} \Id{t}(\=\Ax{1}, \Ax{1}, \Ax{1}) \Ax{=} \Ax{1}\\
\KW{end}\\
\VIEW \=\SId{WeakenedLeft} \Ax{:} \=\SId{Generic} \KW{to} \SId{WeakenedLeftRule}\\
\KW{end}\\
\VIEW \=\SId{WeakenedRight} \Ax{:} \=\SId{Generic} \KW{to} \SId{WeakenedRightRule}\\
\KW{end}\\
\\
\%\% A computed blend as new 8 bit encoding\\
\SPEC \=\SIdIndex{ConsistentBlend} \Ax{=}\\
\> \KW{combine} \=\Id{WeakenedLeft}, \Id{WeakenedRight}\\
\AND \SId{METACABitcalc}\\
\AND \SId{LocalRule}\\
\KW{end}
\end{hetcasl}
}
\caption{CASL source code listing calculating the running example $01101110\times 01010100\times 01010101$ via the blending meta-rule\label{CASL-listing}}
\end{listing}

\newpage
\begin{figure}[H]
\includegraphics[width=0.4\textwidth]{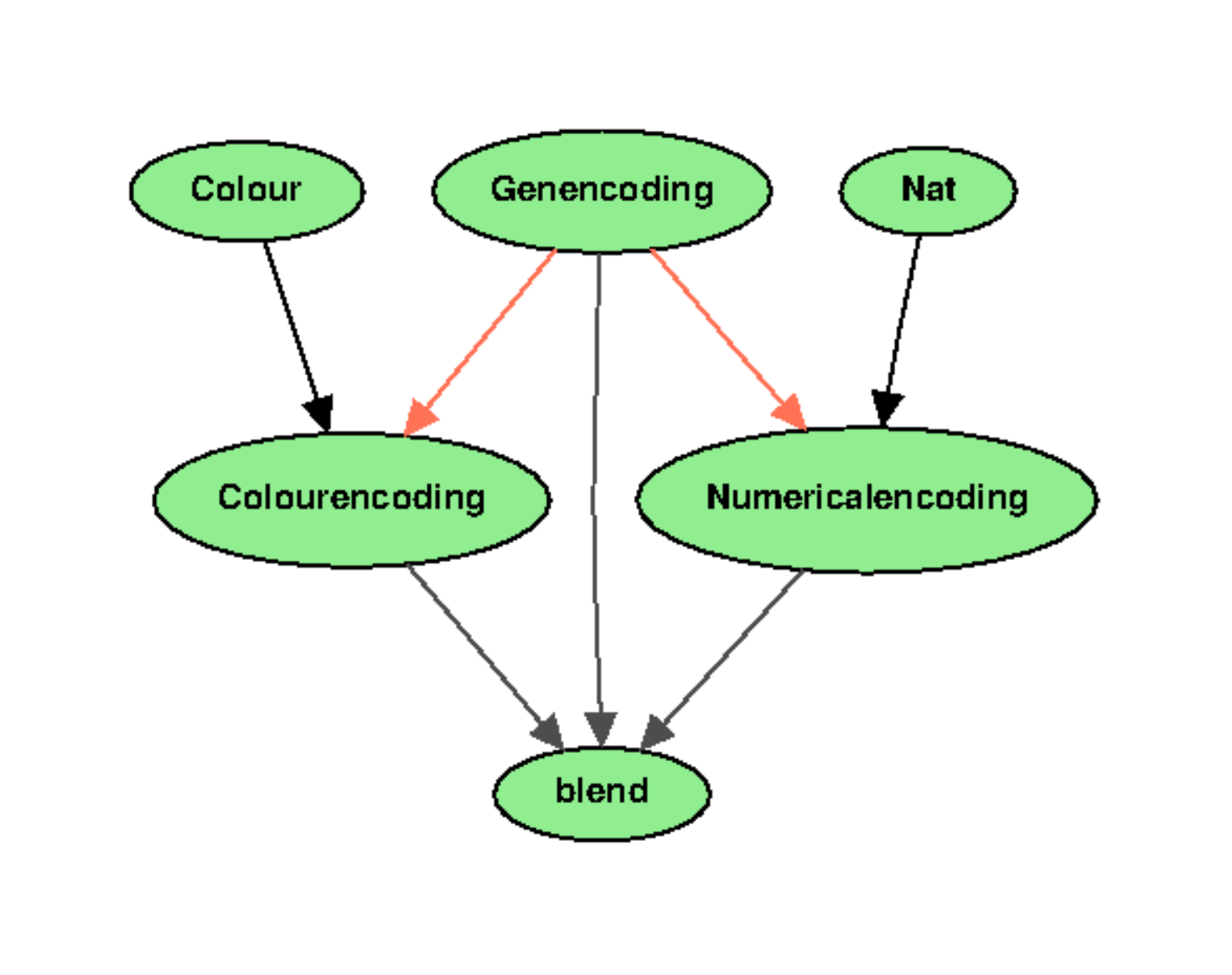}
\caption{Blending different 2d genotypes}
\label{fig:2dblend}
\end{figure}

\begin{listing}[H]
{\scriptsize
\begin{hetcasl}
\KW{library} \Id{metaca2d}\\
\\
\KW{logic} \SId{CASL}\\
\\
\SPEC \=\SIdIndex{Nat} \Ax{=}\\
\> \SORT \Id{Nat}\\
\> \OP \=\Id{max} \Ax{:} \=\Id{Nat} \Ax{\times} \Id{Nat} \Ax{\rightarrow} \Id{Nat}\\
\> \OP \=\Id{min} \Ax{:} \=\Id{Nat} \Ax{\times} \Id{Nat} \Ax{\rightarrow} \Id{Nat}\\
\KW{end}\\
\\
\SPEC \=\SIdIndex{Colour} \Ax{=}\\
\> \SORT \Id{Colour}\\
\> \OP \=\Id{maxhue} \Ax{:} \=\Id{Colour} \Ax{\times} \Id{Colour} \Ax{\rightarrow} \Id{Colour}\\
\KW{end}\\
\\
\\
{{}\KW{\%\%} a 2\Ax{-}d cellular automaton with numerical Genotype}\\
\SPEC \=\SIdIndex{Numericalencoding} \Ax{=}\\
\> \SId{Nat}\\
\THEN \=\SORT \Id{NGenotype}\\
\> \OPS \=\Id{genotype} \Ax{:} \=\Id{Nat} \Ax{\times} \Id{Nat} \Ax{\times} \Id{Nat} \Ax{\rightarrow} \Id{NGenotype};\\
\>\> \Id{t} \Ax{:} \=\Id{Nat} \Ax{\times} \Id{Nat} \Ax{\times} \Id{Nat} \Ax{\rightarrow} \Id{NGenotype};\\
\>\> \Id{numblend} \Ax{:} \=\Id{NGenotype} \Ax{\times} \Id{NGenotype} \Ax{\rightarrow} \Id{NGenotype}\\
\> \Ax{\forall} \Id{g1}, \Id{g2} \Ax{:} \Id{NGenotype}; \=\Id{x1}, \Id{y1}, \Id{z1}, \Id{x2}, \Id{y2}, \Id{z2}, \Id{x3}, \Id{y3}, \Id{z3} \Ax{:} \Id{Nat} \\
\> \Ax{\bullet} \=\Id{g1} \Ax{=} \Id{t}(\=\Id{x1}, \Id{y1}, \Id{z1}) \Ax{\wedge} \=\Id{g2} \Ax{=} \Id{t}(\=\Id{x2}, \Id{y2}, \Id{z2}) \\
\>\> \Ax{\Rightarrow} \=\Id{numblend}(\=\Id{g1}, \Id{g2}) \\
\>\>\> \Ax{=} \Id{t}(\=\Id{min}(\=\Id{x1}, \Id{x2}), \Id{min}(\=\Id{y1}, \Id{y2}), \Id{max}(\=\Id{z1}, \Id{z2}))\\
\KW{end}\\
\\
\\
{{}\KW{\%\%} A colour CA Genotype}\\
\SPEC \=\SIdIndex{Colourencoding} \Ax{=}\\
\> \SId{Colour}\\
\THEN \=\SORT \=\Id{CGenotype} \Ax{=} \Id{Colour}\\
\> \OP \=\Id{hueblend} \Ax{:} \=\Id{CGenotype} \Ax{\times} \Id{CGenotype} \Ax{\rightarrow} \Id{CGenotype}\\
\> \Ax{\forall} \=\Id{g1}, \Id{g2} \Ax{:} \Id{CGenotype} \\
\> \Ax{\bullet} \=\Id{hueblend}(\=\Id{g1}, \Id{g2}) \Ax{=} \Id{maxhue}(\=\Id{g1} \Id{as} \Id{Colour}, \=\Id{g2} \Id{as} \Id{Colour})\\
\KW{end}\\
\\
\\
{{}\KW{\%\%} A generic space}\\
\SPEC \=\SIdIndex{Genencoding} \Ax{=}\\
\> \SORT \Id{S}\\
\> \SORT \Id{Genotype}\\
\> \OP \=\Id{blend} \Ax{:} \=\Id{Genotype} \Ax{\times} \Id{Genotype} \Ax{\rightarrow} \Id{Genotype}\\
\KW{end}\\
\\
\\
{{}\KW{\%\%} A signature morphism from Generic to Numerical}\\
\VIEW \=\SId{NumericalSM} \Ax{:} \\
\> \SId{Genencoding} \KW{to} \SId{Numericalencoding} \Ax{=} \\
\> \Id{S} \Ax{\mapsto} \Id{Nat}, \Id{Genotype} \Ax{\mapsto} \Id{NGenotype}, \=\Id{blend} \Ax{\mapsto} \Id{numblend}\\
\KW{end}\\
\\
\\
{{}\KW{\%\%} A signature morphism from Generic to Colour}\\
\VIEW \=\SId{ColourSM} \Ax{:} \\
\> \SId{Genencoding} \KW{to} \SId{Colourencoding} \Ax{=} \\
\> \Id{S} \Ax{\mapsto} \Id{Colour}, \Id{Genotype} \Ax{\mapsto} \Id{CGenotype}, \=\Id{blend} \Ax{\mapsto} \Id{hueblend}\\
\KW{end}\\
\\
\SPEC \=\SIdIndex{blend} \Ax{=}\\
\> \KW{combine} \=\Id{NumericalSM}, \Id{ColourSM}\\
\KW{end}
\end{hetcasl}

}
\caption{CASL source code using signature morphisms and pushout
  calculation to blend genotypes with different
  languages\label{CASL-blend-listing}}
\end{listing}

\FloatBarrier

\section{Discussion}

\subsection{Research Contribution}

The motivation for combining a notion of blending with cellular
automata was to investigate ways in which cellular automata could be
used to model processes, where propagation rules, or genotypes, were
locally defined.  The main research contributions in the field of two
dimensional cellular automata are
\begin{itemize}
\item We built and implemented a framework where local propagation
  experiments can be performed;
\item We used the HETS system to show that the notion of blending can
  be used to invent new propagation rules for different genotypes;
\item We invented simply definable genotypes and blends of these
  genotypes to show proof of concept;
\item Finally, we shared the results of simulations that illustrate
  qualitative behaviour in one and two dimensional MetaCAs.
\end{itemize}

The primary limitation of this work is that our results are purely
observational at present.  For example, the early experiments seemed
to provide visual evidence that blending is useful: Figure
\ref{metaca-taster} more interesting than Figure \ref{barcode}.  The
robustness of our qualitative findings have been supported by
developing a range of different experiments, for example, some analogy
could be drawn between the ``grey areas'' observed in Figure
\ref{seti} for the 1D case and the red-and-white chains that develop
in the 2D case under union blending.

Our results confirm the basic finding of CA research: interesting
global behaviour can arise from simple rules governing local
interactions, with the added twist the rules can also arise locally.
The MetaCA setting seems to offer fertile ground for further
computational research into evolutionary and co-evolutionary effects.

\subsection{Social Interpretation}

One can view the propagation of cells and patterns in a 1D or 2D
MetaCA as a social process, and blending as a knowledge exchange.  In
the 2D case, we can think of the generated diagrams as illustrations
of interactions between individuals with high knowledge, skill, or
social impact (high weight), and those with less (low weight).  The
propagation in the ``union'' blend shows how large numbers of
individuals with low social impact outnumber those with high social
impact, but those with high social impact impose the emergent
structure and determine the growth of the group of individuals.

In a fundamental respect our blending rules seem to embody a
thought-provoking blend of two very different kinds of ``ethics.''
Specifically: blending seems to introduce a dynamic similar to Carol
Gilligan's \emph{ethic of care} \cite{gilligan1982different}, which
seeks to defend the relationships that obtain in a given situation.
Here this is manifested by the question ``Have my neighbours already
formed a consensus?''  This behaviour augments and extends the local
rule, which would correspond to Lawrence Kohlberg’s \emph{ethic of
  justice} (cf. \cite{benhabib1985generalized}).

As we saw in Section \ref{1d-results}, we would have to work harder to
find meta-rules that give rise to an ``intelligent universe'' or in
which life (considered as symbolic computation) plays an obvious
negentropic role (\emph{apr\`es} Bergson \cite{bergson1912creative}).

One strategy that has not been developed here would be to make use of
a ``Baldwin effect'' \cite{baldwin-effect,weber2003evolution}, to use
``learning'' (considered as entropy) in the phenotype layer to drive
evolution.  More specifically,
\boxed{0\mystrut}\boxed{0\mystrut}\boxed{0\mystrut} $\mapsto$
\boxed{0\mystrut} and
\boxed{1\mystrut}\boxed{1\mystrut}\boxed{1\mystrut} $\mapsto$
\boxed{1\mystrut} seem to be relatively uninteresting behaviours, but
they are also hard to resist under the blending dynamics as we've
defined them (compare Figures \ref{flag} and \ref{seti}).  Can we find
ways to select against them?

\subsection{Planned extensions}

One observes that under our blending rule, the two non-entropic
behaviours listed above tend to selected for, not against, because
they are examples of the ``neighbours match'' condition.  Indeed,
reviewing the essential features of blending in the 1D case, we can
use our basic principles:

\begin{quote}
``\emph{If neighbours match:} \emph{use their shared value as the result.}\\
 \emph{If neighbours don't match:} \emph{use local logic to get the result.}''
\end{quote}

\noindent to define a 1D CA rule, if we interpret ``local logic'' to
mean ``substitute my own value as the result.''  Here's how we would
then define blending for triplets:

\lstset{
  xleftmargin=.2\columnwidth, xrightmargin=.01\columnwidth
}

\begin{lstlisting}[mathescape]
0 0 0 $\mapsto$ 0    $\text{\emph{Neighbours match}}$
0 0 1 $\mapsto$ 0    $\text{\emph{Local logic}}$
0 1 0 $\mapsto$ 0    $\text{\emph{Neighbours match}}$
0 1 1 $\mapsto$ 1    $\text{\emph{Local logic}}$
1 0 0 $\mapsto$ 0    $\text{\emph{Local logic}}$
1 0 1 $\mapsto$ 1    $\text{\emph{Neighbours match}}$
1 1 0 $\mapsto$ 1    $\text{\emph{Local logic}}$
1 1 1 $\mapsto$ 1    $\text{\emph{Neighbours match}}$
\end{lstlisting}

This is Wolfram's Rule 23: and as it happens, its evolutionary
behaviour is not particularly interesting.  Of course, for blending at
the genotype level, ``local logic'' can be determined by any CA.  Even
so, when we use blending bitwise on alleles, we only ever run the
local logic on half of the cases, and moreover it always the same
half, determined by a ``censored'' version of Rule 23.

\begin{lstlisting}[mathescape]
0 0 0 $\mapsto$ 0    $\text{\emph{Neighbours match}}$
0 0 1 $\mapsto$ *    $\text{\emph{Local logic}}$
0 1 0 $\mapsto$ 0    $\text{\emph{Neighbours match}}$
0 1 1 $\mapsto$ *    $\text{\emph{Local logic}}$
1 0 0 $\mapsto$ *    $\text{\emph{Local logic}}$
1 0 1 $\mapsto$ 1    $\text{\emph{Neighbours match}}$
1 1 0 $\mapsto$ *    $\text{\emph{Local logic}}$
1 1 1 $\mapsto$ 1    $\text{\emph{Neighbours match}}$
\end{lstlisting}

Rather than using Censored Rule 23 as our template, we could instead
have the template determined by phenotype data, thereby inserting the
phenotype as a ``hidden layer'' in the computation.

The standard template could be understood to be generated by locking in
\boxed{0\mystrut}\boxed{0\mystrut}\boxed{0\mystrut} $\mapsto$ \boxed{0\mystrut}
along with a ``variation''\footnote{%
\boxed{0\mystrut}\boxed{1\mystrut}\boxed{0\mystrut} =
\boxed{0\mystrut}\boxed{0\mystrut}\boxed{0\mystrut} +
\boxed{\:\:\:\mystrut}\boxed{1\mystrut}\boxed{\:\:\:\mystrut}}
\boxed{0\mystrut}\boxed{1\mystrut}\boxed{0\mystrut} $\mapsto$
\boxed{0\mystrut} and the bitwise inverses of these.  A wider class of
templates could calculated from arbitrary phenotype data by the same
operations.  What we would lose in abandoning the intuition associated
with local blending, we may be repaid through a much more abstract
but richer procedural blend, operating at the level of
genotype+phenotype evolution.  At the very least, we can point to a
generic space, namely the locked-in local rule which would be carried
over (along with its variants) from the phenotype to the corresponding
alleles.

As a simple example of cross-domain blending consider a genotype
defined as in \S\ref{2d-experiments-design}, and another which is
defined by comparing the hue of just one neighbour.  Their blend is a
richer theory combining elements from both genotypes.  CASL code
expressing these concepts is given in Listing
\ref{CASL-blend-listing}, and the resulting categorical diagram can be
seen in Figure \ref{fig:2dblend}.  Experimentation with more
sophisticated genotypes and blends is ongoing.

\subsection{Future work}

Coevolution has been understood to be relevant from both a
philosophical \cite{mead1932philosophy} and empirical perspective
\cite{van1973new}.  Finding patterns that allow us to exploit Baldwin
effects to drive the co-evolution of genotype and phenotype in the
direction of intelligent behaviour is an interesting computational
project.  The MetaCA domain may help to show how to systematise some
aspects of the search for the principles and techniques that underlie
broader computational intelligence.

Expanding on the relatively simple domain of CAs, we would like to use
HETS to formalise the mechanisms of social knowledge sharing and
problem solving in fields like mathematics.  Indeed it may be possible
in the future to encode mathematical problems in a MetaCA or cellular
program and see how a group of agents can solve the problems as a
society.  This would be informed by ongoing empirical analysis of real
problem-solving activities \cite{eca} developed in parallel to the
simulation work presented here.

\section{Conclusion}

This research was inspired by the aim to build an example of
computational blending that matched, to some extent, the way blending
might work in social settings.  One person suggests an idea, and
another offers a variant of that, a third brings in another idea from
elsewhere and some combination is made.  The next day, things head in
another direction completely.  Our progress in this research project
has followed this sort of trajectory: from an initial critique of
blending theory (``it's not dynamic enough to be social!'') to some
tentative examples showing how large-scale system dynamics can be
driven by local behaviour in an emergent manner.  Perhaps the most
interesting aspect of this research is the relationship between these
emergent dynamics and the meta-rules.  Whereas previous CA research
has shown that complex global behaviour can be generated from a set of
simple, local rules, this project gives an enticing glimpse of a
future research programme that carries out a computational search for
those very rules (out of the many possible) that lead to system
behaviour we would recognise as ``intelligent.''

\section{Acknowledgements}

This research has been funded by the Future and Emerging Technologies
(FET) programme within the Seventh Framework Programme for Research of
the European Commission, under FET-Open Grant number 611553
(COINVENT).

We thank Timothy Teravainen, Raymond Puzio, and Cameron Smith for
helpful conversations and pointers to literature.

\bibliography{metaca}

\end{document}